\documentclass[preprint,12pt, a4paper]{elsarticle}

\usepackage{amssymb}
\usepackage{hyperref}
\usepackage{comment}
\usepackage{caption} 
\usepackage[utf8]{inputenc} 
\usepackage[T1]{fontenc}    
\usepackage{times}  
\usepackage{helvet} 
\usepackage{courier}  
\usepackage{booktabs}       
\usepackage{nicefrac}       
\usepackage{microtype}      
\usepackage{caption}
\usepackage{listings}
\usepackage{multirow}
\lstdefinestyle{mystyle}{
        language=python,
        basicstyle=\footnotesize\ttfamily,
        breaklines=true,
        keepspaces=true,
    }
\usepackage{adjustbox}
\usepackage{sidecap}
\usepackage{subcaption}
\usepackage{makecell}
\usepackage{amsmath}
\usepackage{lineno}
\usepackage{enumitem}
\usepackage{comment}
\usepackage{array}
\newcolumntype{L}{>{\centering\arraybackslash}m{3cm}}

\journal{Software Impacts}

\begin{document}

\begin{frontmatter}



\title{Spark NLP: Natural Language Understanding at Scale}


\author{Veysel Kocaman,
    David Talby}

\address{John Snow Labs Inc.\\


    16192 Coastal Highway\\
    Lewes, DE , USA 19958\\
    \{veysel, david\}@johnsnowlabs.com}

\begin{abstract}


Spark NLP is a Natural Language Processing (NLP) library built on top of Apache Spark ML. It provides simple, performant \& accurate NLP annotations for machine learning pipelines that can scale easily in a distributed environment. Spark NLP comes with 1100+ pretrained pipelines and models in more than 192+ languages. It supports nearly all the NLP tasks and modules that can be used seamlessly in a cluster. Downloaded more than 2.7 million times and experiencing 9x growth since January 2020, Spark NLP is used by 54\% of healthcare organizations as the world’s most widely used NLP library in the enterprise.


\end{abstract}


\end{frontmatter}


\noindent
\section{Spark NLP Library}
\label{sec:intro}
Natural language processing (NLP) is a key component in many data science systems that must understand or reason about a text. Common use cases include question answering, paraphrasing or summarising, sentiment analysis, natural language BI, language modelling, and disambiguation. Nevertheless, NLP  is always just a part of a bigger data processing pipeline and due to the nontrivial steps involved in this process, there is a growing need for all-in-one solution to ease the burden of text preprocessing at large scale and connecting the dots between various steps of solving a data science problem with NLP. A good NLP library should be able to correctly transform the free text into structured features and let the users train their own NLP models that are easily fed into the downstream machine learning (ML) or deep learning (DL) pipelines with no hassle.

Spark NLP is developed to be a single unified solution for all the NLP tasks and is the only library that can scale up for training and inference in any Spark cluster, take advantage of transfer learning and implementing the latest and greatest algorithms and models in NLP research, and deliver a mission-critical, enterprise-grade solutions at the same time. It is an open-source natural language processing library, built on top of Apache Spark and Spark ML. It provides an easy API to integrate with ML pipelines and it is commercially supported by John Snow Labs Inc, an award-winning healthcare AI and NLP company based in USA. 

Spark NLP’s annotators utilize rule-based algorithms, machine learning and deep learning models which are implemented using TensorFlow that has been heavily optimized for accuracy, speed, scalability, and memory utilization. This setup has been tightly integrated with Apache Spark to let the driver node run the entire training using all the available cores on the driver node. There is a CuDA version of each TensorFlow component to enable training models on GPU when available. The Spark NLP is written in Scala and provides open-source API's in Python, Java, Scala, and R - so that users do not need to be aware of the underlying implementation details (TensorFlow, Spark, etc.) in order to use it.  Since it has an active release cycle (released 26 new versions in 2019 and another 26 in 2020), the latest trends and research in NLP field are embraced and implemented rapidly in a way that could scale well in a cluster setting to allow common NLP pipelines run orders of magnitude faster than what the inherent design limitations of legacy libraries allowed.

\begin{figure*}[htb]
\includegraphics[width=1.0\textwidth,scale=1.0]{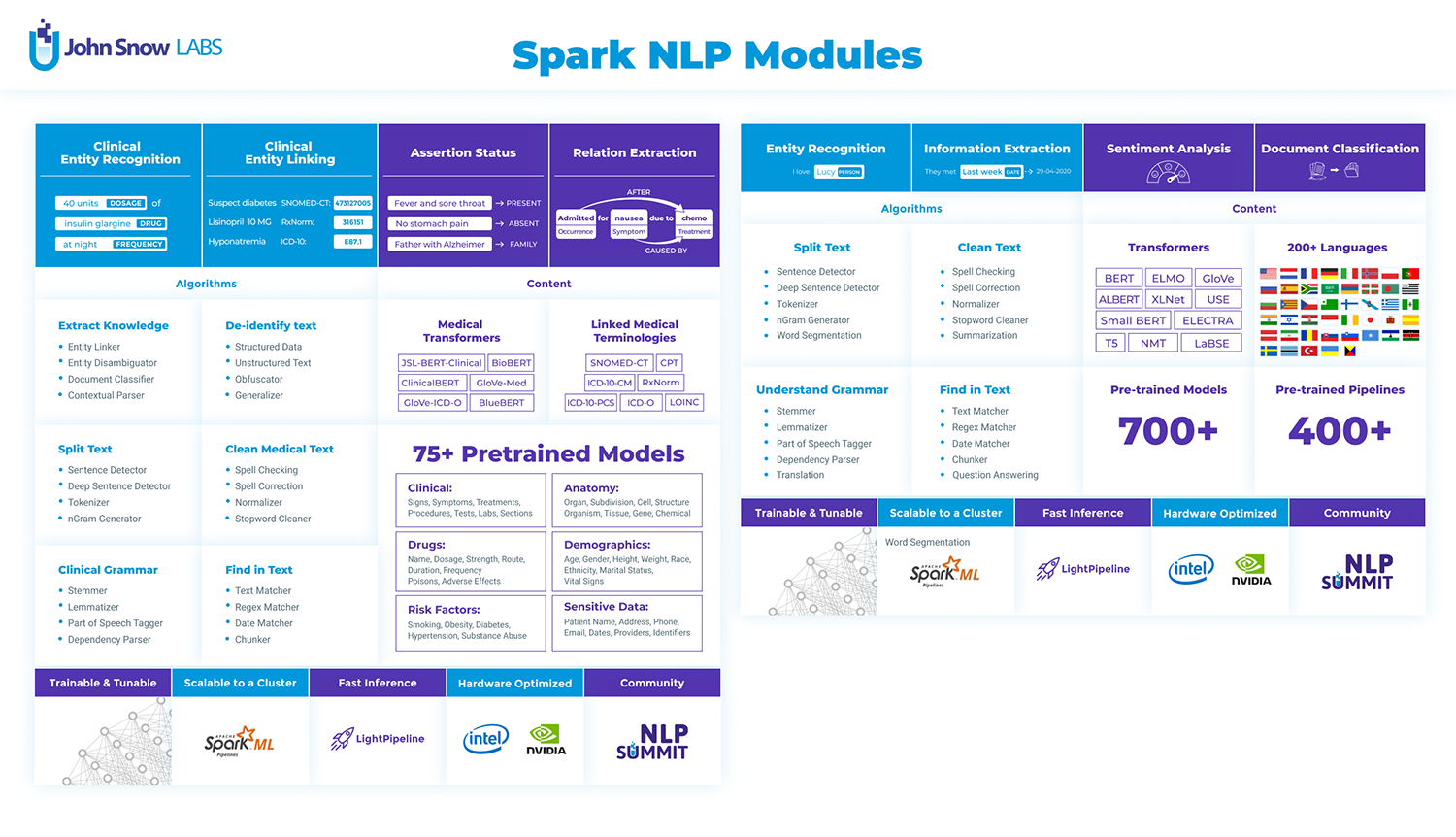}
\centering
\caption{Spark NLP library has two versions (open source and enterprise) and each comes with a set of pretrained models and pipelines that could be used out of the box with no further training or dataset.}
\label{fig:spark_modules}
\end{figure*}

Spark NLP library has two versions: Open source and enterprise. Open source version has all the features and components that could be expected from any NLP library, using the latest DL frameworks and research trends. Enterprise library is licensed  (free for academic purposes) and designed towards solving real world problems in healthcare domain and extends the open source version. The licensed version has the following modules to help researchers and data practitioners in various means: Named entity recognition (NER), assertion status (negativity scope) detection, relation extraction, entity resolution (SNOMED, RxNorm, ICD10 etc.), clinical spell checking, contextual parser, text2SQL, deidentification and obfuscation. High level overview of the components from each version can be seen at Figure~\ref{fig:spark_modules}.

\section{The impact to research fields}
\label{sec:impact}

The COVID-19 pandemic brought a surge of academic research about the virus - resulting in 23,634 new publications between January and June of 2020 \cite{Silva2020PublishingVI} and accelerating to 8,800 additions per week from June to November on the COVID-19 Open Research Dataset \cite{wang2020cord}. Such a high volume of publications makes it impossible for researchers to read each publication, resulting in increased interest in applying natural language processing (NLP) and text mining techniques to enable semi-automated literature review \cite{Cheng2020AnOO}.

In parallel, there is a growing need for automated text mining of Electronic health records (EHRs) in order to find clinical indications that new research points to. EHRs are the primary source of information for clinicians tracking the care of their patients. Information fed into these systems may be found in structured fields for which values are inputted electronically (e.g. laboratory test orders or results)~\cite{liede2015validation} but most of the time information in these records is unstructured making it largely inaccessible for statistical analysis~\cite{murdoch2013inevitable}. These records include information such as the reason for administering drugs, previous disorders of the patient or the outcome of past treatments, and they are the largest source of empirical data in biomedical research, allowing for major scientific findings in highly relevant disorders such as cancer and Alzheimer’s disease ~\cite{perera2014factors}. Despite the growing interest and ground breaking advances in NLP research and NER systems, easy to use production ready models and tools are scarce in biomedical and clinical domain and it is one of the major obstacles for clinical NLP researchers to implement the latest algorithms into their workflow and start using immediately. On the other hand, NLP tool kits specialized for processing biomedical and clinical text, such as MetaMap~\cite{aronson2010overview} and cTAKES~\cite{savova2010mayo} typically do not make use of new research innovations such as word representations or neural networks discussed above, hence producing less accurate results~\cite{zhang2020biomedical, neumann2019scispacy}. We introduce Spark NLP as the one-stop solution to address all these issues.

\begin{figure*}[htb]
\includegraphics[width=0.9\textwidth,scale=0.9]{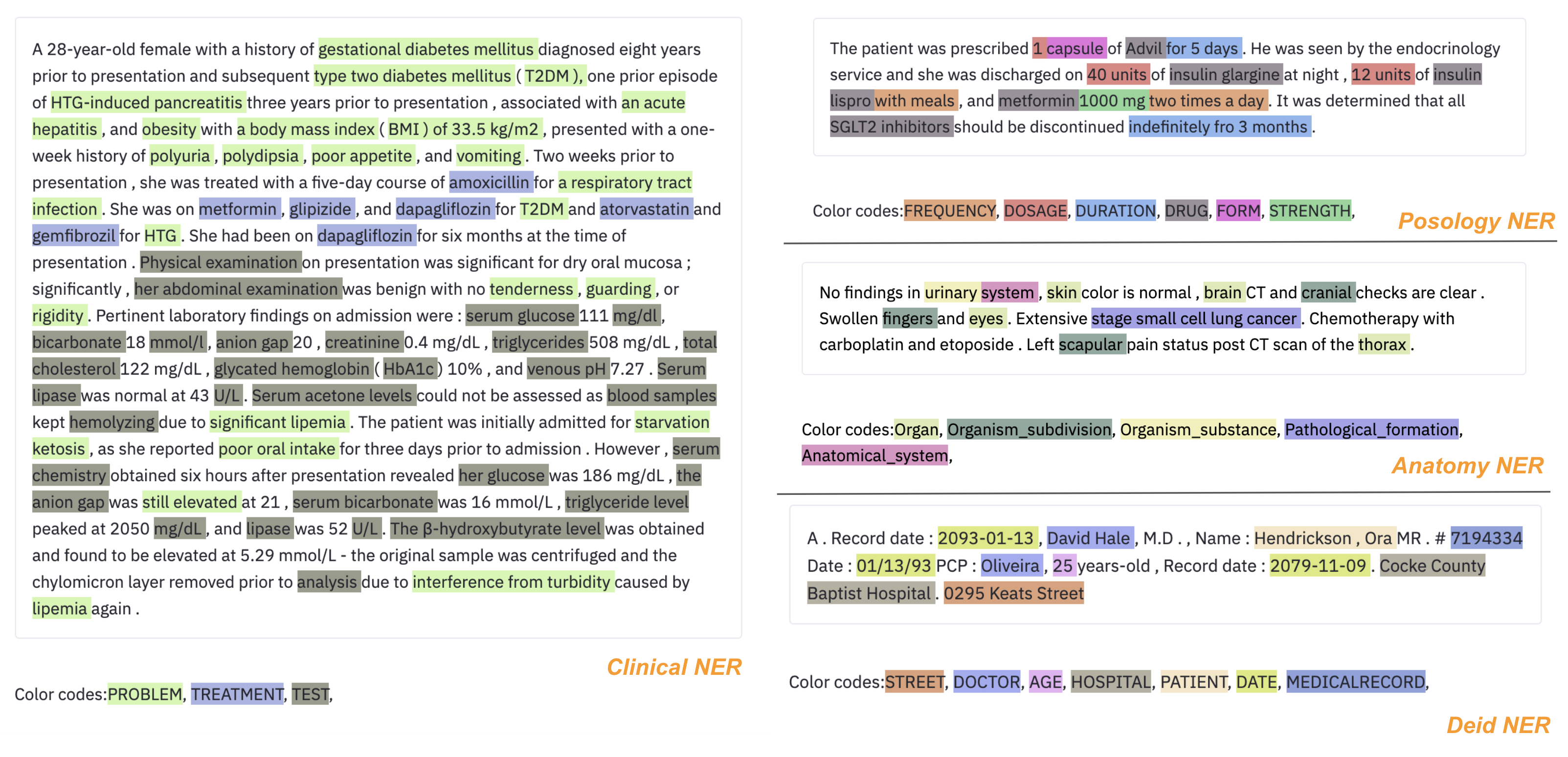}
\centering
\caption{Sample clinical entities predicted by a clinical NER model trained on various datasets. There are more than 40 pretrained NER models in Spark NLP Enterprise edition.}
\label{fig:clinical_ner}
\end{figure*}

A primary building block in such text mining systems is named entity recognition (NER) - which is regarded as a critical precursor for question answering, topic modelling, information retrieval, etc~\cite{yadav2019survey}.  In the medical domain, NER recognizes the first meaningful chunks out of a clinical note, which are then fed down the processing pipeline as an input to subsequent downstream tasks such as clinical assertion status detection~\cite{uzuner20112010}, clinical entity resolution ~\cite{tzitzivacos2007international} and de-identification of sensitive data~\cite{uzuner2007evaluating}. However, segmentation of clinical and drug entities is considered to be a difficult task in biomedical NER systems because of complex orthographic structures of named entities ~\cite{liu2015effects}. Sample NER predictions from a clinical text can be found at Figure~\ref{fig:clinical_ner}. 

\begin{figure*}[htb]
\includegraphics[width=1.0\textwidth, scale=0.7]{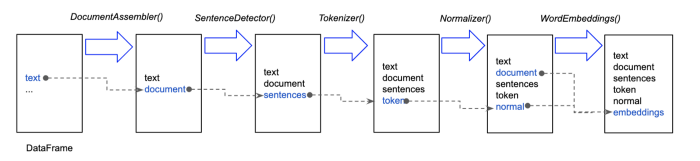}
\centering
\caption{The flow diagram of a Spark NLP pipeline. When we fit() on the pipeline with a Spark data frame, its text column is fed into the DocumentAssembler() transformer and a new column \textit{document} is created as an initial entry point to Spark NLP for any Spark data frame. Then, its document column is fed into the SentenceDetector() module to split the text into an array of sentences and a new column “sentences” is created. Then, the “sentences” column is fed into Tokenizer(), each sentence is tokenized, and a new column “token” is created. Then, Tokens are normalized (basic text cleaning) and word embeddings are generated for each. Now data is ready to be fed into NER models and then to the assertion model. }
\label{fig:pipeline_diagram}
\end{figure*}

The next step following an NER model in the clinical NLP pipeline is to assign an assertion status to each named entity given its context. The status of an assertion explains how a named entity (e.g. clinical finding, procedure, lab result) pertains to the patient by assigning a label such as present ("patient is diabetic"), absent ("patient denies nausea"), conditional ("dyspnea while climbing stairs"), or associated with someone else ("family history of depression"). In the context of COVID-19, applying an accurate assertion status detection is crucial, since most patients will be tested for and asked about the same set of symptoms and comorbidities - so limiting a text mining pipeline to recognizing medical terms without context is not useful in practice. The flow diagram of such a pipeline can be seen in Figure~\ref{fig:pipeline_diagram}. 

In our previous study~\cite{kocaman2020biomedical}, we showed through extensive experiments that NER module in Spark NLP library exceeds the biomedical NER benchmarks reported by Stanza in 7 out of 8 benchmark datasets and in every dataset reported by SciSpacy without using heavy contextual embeddings like BERT. Using the modified version of the well known BiLSTM-CNN-Char NER architecture~\cite{chiu2016named} into Spark environment, we also presented that even with a general purpose GloVe embeddings (GloVe6B) and with no lexical features, we were able to achieve state-of-the-art results in biomedical domain and produces better results than Stanza in 4 out of 8 benchmark datasets. 

\begin{figure*}[htb]
\includegraphics[width=0.9\textwidth,scale=0.9]{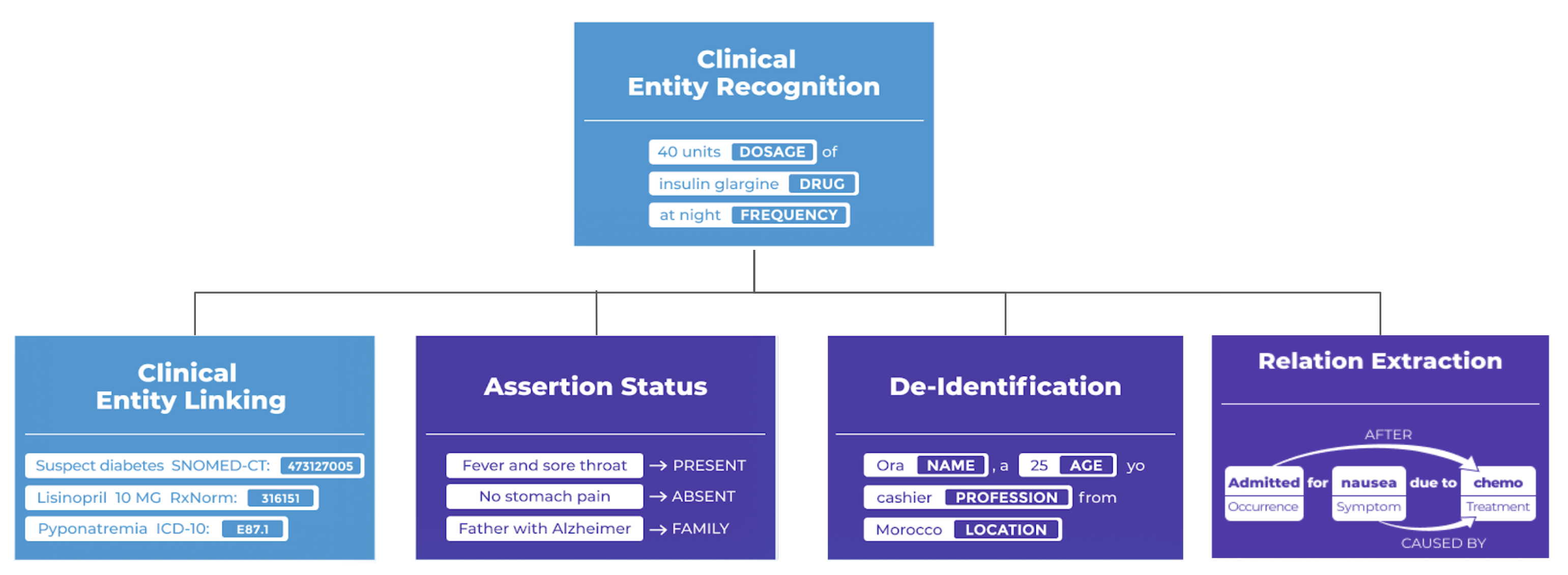}
\centering
\caption{Named Entity Recognition is a fundamental building block of medical text mining pipelines, and feeds downstream tasks such as assertion status, entity linking, de-identification, and relation extraction.}
\label{fig:ner_tree_chart}
\end{figure*}

In another study~\cite{kocaman2020improving}, we introduced a set of pre-trained NER models that are all trained on biomedical and clinical datasets using the same deep learning architecture. We then illustrated how to extract knowledge and relevant information from unstructured electronic health records (EHR) and COVID-19 Open Research Dataset (CORD-19) by combining these models in a unified \& scalable pipeline and shared the results to illustrate extracting valuable information from scientific papers. The results suggest that papers present in the CORD-19 include a wide variety of the many entity types that this new NLP pipeline can recognize, and that assertion status detection is a useful filter on these entities (Figure~\ref{fig:ner_tree_chart}). The most frequent phrases from the selected entity types can be found at Table~\ref{tab:topterms}. This bodes well for the richness of downstream analysis that can be done using this now structured and normalized data - such as clustering, dimensionality reduction, semantic similarity, visualization, or graph-based analysis to identity correlated concepts. Moreover, in order to evaluate how fast the pipeline works and how effectively it scales to make use of a compute cluster, we ran the same Spark NLP prediction pipelines in local mode and in cluster mode: and found out that tokenization is 20x faster while the entity extraction is 3.5x faster on the cluster, compared to the single machine run. 

\section{The impact to industrial and academic collaborations}
\label{sec:coll}

As the creator of Spark NLP, John Snow Labs company has been supporting the researchers around the globe by distributing them a free license to use all the licensed modules both in research projects and graduate level courses at universities, providing hands-on supports when needed, organizing workshops and summits to gather distinguished speakers and running projects with the R\&D teams of the top pharmacy companies to help them unlock the potential of unstructured text data buried in their ecosystem. Spark NLP already powers leading healthcare and pharmaceutical companies including Kaiser Permanente, McKesson, Merck, and Roche. Since Spark NLP can also be used offline and deployed in air-gapped networks, the companies and healthcare facilities do not need to worry about exposing the protected health information (PHI). The detailed information about these projects and case studies can be found at ~\cite{lessonslearned}, ~\cite{jslnlpcasestudies}, ~\cite{jslcasestudies}.

\begin{table*}[htb]
\caption{NER performance across different datasets in the biomedical domain. All scores reported
are micro-averaged test F1 excluding O's. Stanza results are from the paper reported in ~\cite{zhang2020biomedical}, SciSpaCy results are from the scispacy-medium models reported in ~\cite{neumann2019scispacy}. The official training and validation sets are merged and used for training and then the models are evaluated on the original test sets.  For reproducibility purposes, we use the preprocessed versions of these datasets provided by ~\cite{wang2019cross} and also used by Stanza. Spark-x prefix in the table indicates our implementation. Bold scores represent the best scores in the respective row.}
\centering
\label{tab:benchmarks}
\resizebox{0.97\textwidth}{!}{
\begin{tabular}{llLLll}
\toprule
Dataset  & Entities   & Spark - Biomedical & Spark - GloVe 6B & Stanza & SciSpacy \\ \midrule
NCBI-Disease & Disease                                 & \textbf{89.13}                       & 87.19                     & 87.49  & 81.65    \\
                     BC5CDR       & Chemical, Disease                       & \textbf{89.73}                       & 88.32                     & 88.08  & 83.92    \\
                     BC4CHEMD     & Chemical                                & \textbf{93.72}                       & 92.32                     & 89.65  & 84.55    \\
                     Linnaeus     & Species                                 & 86.26                       & 85.51                     & \textbf{88.27}  & 81.74    \\
                     Species800   & Species                                 & \textbf{80.91}                       & 79.22                     & 76.35  & 74.06    \\
                     JNLPBA       & 5 types in cellular & \textbf{81.29}                       & 79.78                     & 76.09  & 73.21    \\
                     AnatEM       & Anatomy                                 & \textbf{89.13}                       & 87.74                     & 88.18  & 84.14    \\
                     BioNLP13-CG  & 16 types in Cancer Genetics & \textbf{85.58}  & 84.30 & 84.34  & 77.60     \\ \bottomrule
\end{tabular}
}
\end{table*}

\begin{table*}[htb]
\caption{The most frequent 10 terms from the selected entity types predicted through parsing 100 articles from CORD-19 dataset~\cite{wang2020cord} with an NER model named \textit{jsl\_ner\_wip} in Spark NLP. Getting predictions from the model, we can get some valuable information regarding the most frequent disorders or symptoms mentioned in the papers or the most common vital and EKG findings without reading the paper. According to this table, the most common symptom is \textit{cough} and \textit{inflammation} while the most common drug ingredients mentioned is \textit{oseltamivir} and \textit{antibiotics}. We can also say that \textit{cardiogenic oscillations} and \textit{ventricular fibrillation} are the common observations from EKGs while \textit{fever} and \textit{hyphothermia} are the most common vital signs.}
\label{tab:topterms} 
\resizebox{0.98\textwidth}{!}{
\begin{tabular}{lllllll}
\hline
\thead{Disease Syndrome\\ Disorder} & \thead{Communicable\\ Disease} & Symptom & \thead{Drug\\ Ingredient} & Procedure & \thead{Vital Sign\\ Findings} & \thead{EKG\\ Findings} \\\hline
infectious diseases & HIV & cough & oseltamivir & resuscitation & fever & low VT \\
sepsis & H1N1 & inflammation & biological agents & cardiac surgery & hypothermia & cardiogenic oscillations \\
influenza & tuberculosis & critically ill & VLPs & tracheostomy & hypoxia & significant changes \\
septic shock & influenza & necrosis & antibiotics & CPR & respiratory failure & CO reduces oxygen transport \\
asthma & TB & bleeding & saline & vaccination & hypotension & ventricular fibrillation \\
pneumonia & hepatitis viruses & lesion & antiviral & bronchoscopy & hypercapnia & significant impedance increases \\
COPD & measles & cell swelling & quercetin & intubation & tachypnea & ventricular fibrillation \\
gastroenteritis & pandemic influenza & hemorrhage & NaCl & transfection & respiratory distress & pulseless electrical activity \\
viral infections & seasonal influenza & diarrhea & ribavirin & bronchoalveolar lavage & hypoxaemia & mildmoderate hypothermia \\
SARS & rabies & toxicity & Norwalk agent & autopsy & pyrexia & cardiogenic oscillations\\
\hline
\end{tabular}
}
\end{table*}

\section{Acknowledgements}
\label{ack}

We thank our colleagues and research partners who contributed in the former and current developments of Spark NLP library. We also thank our users and customers who helped us improve the library with their feedbacks and suggestions.



\bibliographystyle{elsarticle-num} 
\bibliography{references.bib}

\begin{thebibliography}{10}
\expandafter\ifx\csname url\endcsname\relax
  \def\url#1{\texttt{#1}}\fi
\expandafter\ifx\csname urlprefix\endcsname\relax\def\urlprefix{URL }\fi
\expandafter\ifx\csname href\endcsname\relax
  \def\href#1#2{#2} \def\path#1{#1}\fi

\bibitem{Silva2020PublishingVI}
J.~A.~T. da~Silva, P.~Tsigaris, M.~Erfanmanesh, Publishing volumes in major
  databases related to covid-19, Scientometrics (2020) 1 -- 12.

\bibitem{wang2020cord}
L.~L. Wang, K.~Lo, Y.~Chandrasekhar, R.~Reas, J.~Yang, D.~Eide, K.~Funk,
  R.~Kinney, Z.~Liu, W.~Merrill, et~al., Cord-19: The covid-19 open research
  dataset, ArXiv.

\bibitem{Cheng2020AnOO}
X.~Cheng, Q.~Cao, S.~Liao, An overview of literature on covid-19, mers and
  sars: Using text mining and latent dirichlet allocation, Journal of
  Information Science.

\bibitem{liede2015validation}
A.~Liede, R.~K. Hernandez, M.~Roth, G.~Calkins, K.~Larrabee, L.~Nicacio,
  Validation of international classification of diseases coding for bone
  metastases in electronic health records using technology-enabled abstraction,
  Clinical epidemiology 7 (2015) 441.

\bibitem{murdoch2013inevitable}
T.~B. Murdoch, A.~S. Detsky, The inevitable application of big data to health
  care, Jama 309~(13) (2013) 1351--1352.

\bibitem{perera2014factors}
G.~Perera, M.~Khondoker, M.~Broadbent, G.~Breen, R.~Stewart, Factors associated
  with response to acetylcholinesterase inhibition in dementia: a cohort study
  from a secondary mental health care case register in london, PloS one 9~(11)
  (2014) e109484.

\bibitem{aronson2010overview}
A.~R. Aronson, F.-M. Lang, An overview of metamap: historical perspective and
  recent advances, Journal of the American Medical Informatics Association
  17~(3) (2010) 229--236.

\bibitem{savova2010mayo}
G.~K. Savova, J.~J. Masanz, P.~V. Ogren, J.~Zheng, S.~Sohn, K.~C.
  Kipper-Schuler, C.~G. Chute, Mayo clinical text analysis and knowledge
  extraction system (ctakes): architecture, component evaluation and
  applications, Journal of the American Medical Informatics Association 17~(5)
  (2010) 507--513.

\bibitem{zhang2020biomedical}
Y.~Zhang, Y.~Zhang, P.~Qi, C.~D. Manning, C.~P. Langlotz, Biomedical and
  clinical english model packages in the stanza python nlp library, arXiv
  preprint arXiv:2007.14640.

\bibitem{neumann2019scispacy}
M.~Neumann, D.~King, I.~Beltagy, W.~Ammar, Scispacy: Fast and robust models for
  biomedical natural language processing, arXiv preprint arXiv:1902.07669.

\bibitem{yadav2019survey}
V.~Yadav, S.~Bethard, A survey on recent advances in named entity recognition
  from deep learning models, arXiv preprint arXiv:1910.11470.

\bibitem{uzuner20112010}
{\"O}.~Uzuner, B.~R. South, S.~Shen, S.~L. DuVall, 2010 i2b2/va challenge on
  concepts, assertions, and relations in clinical text, Journal of the American
  Medical Informatics Association 18~(5) (2011) 552--556.

\bibitem{tzitzivacos2007international}
D.~Tzitzivacos, International classification of diseases 10th edition
  (icd-10):: main article, CME: Your SA Journal of CPD 25~(1) (2007) 8--10.

\bibitem{uzuner2007evaluating}
{\"O}.~Uzuner, Y.~Luo, P.~Szolovits, Evaluating the state-of-the-art in
  automatic de-identification, Journal of the American Medical Informatics
  Association 14~(5) (2007) 550--563.

\bibitem{liu2015effects}
S.~Liu, B.~Tang, Q.~Chen, X.~Wang, Effects of semantic features on machine
  learning-based drug name recognition systems: word embeddings vs. manually
  constructed dictionaries, Information 6~(4) (2015) 848--865.

\bibitem{kocaman2020biomedical}
V.~Kocaman, D.~Talby, Biomedical named entity recognition at scale, arXiv
  preprint arXiv:2011.06315.

\bibitem{chiu2016named}
J.~P. Chiu, E.~Nichols, Named entity recognition with bidirectional lstm-cnns,
  Transactions of the Association for Computational Linguistics 4 (2016)
  357--370.

\bibitem{kocaman2020improving}
V.~Kocaman, D.~Talby, Improving clinical document understanding on covid-19
  research with spark nlp, arXiv preprint arXiv:2012.04005.

\bibitem{lessonslearned}
J.~S. Labs, {Apache Spark NLP for Healthcare: Lessons Learned Building
  Real-World Healthcare AI Systems},
  \url{https://databricks.com/session_na20/apache-spark-nlp-for-healthcare-lessons-learned-building-real-world-healthcare-ai-systems},
  [Online; accessed 22-Jan-2021] (2021).

\bibitem{jslnlpcasestudies}
J.~S. Labs, {NLP Case Studies},
  \url{https://www.johnsnowlabs.com/nlp-case-studies/}, [Online; accessed
  22-Jan-2021] (2021).

\bibitem{jslcasestudies}
J.~S. Labs, {AI Case Studies},
  \url{https://www.johnsnowlabs.com/ai-case-studies/}, [Online; accessed
  22-Jan-2021] (2021).

\bibitem{wang2019cross}
X.~Wang, Y.~Zhang, X.~Ren, Y.~Zhang, M.~Zitnik, J.~Shang, C.~Langlotz, J.~Han,
  Cross-type biomedical named entity recognition with deep multi-task learning,
  Bioinformatics 35~(10) (2019) 1745--1752.

\end{thebibliography}






\section*{Required Metadata}
\label{}

\section*{Current code version}
\label{meta1}
\begin{table}[!h]
\begin{tabular}{|l|p{6.5cm}|p{6.5cm}|}
\hline
\textbf{Nr.} & \textbf{Code metadata description} & \textbf{Please fill in this column} \\
\hline
C1 & Current code version & v2.7.1 \\
\hline
C2 & Permanent link to code/repository used for this code version & https://github.com/JohnSnowLabs/spark-nlp \\
\hline
C3  & Permanent link to Reproducible Capsule & https://github.com/JohnSnowLabs/spark-nlp-workshop \\
\hline
C4 & Legal Code License   &  Apache-2.0 License \\
\hline
C5 & Code versioning system used & git, maven \\
\hline
C6 & Software code languages, tools, and services used & scala, python, java, R \\
\hline
C7 & Compilation requirements, operating environments \& dependencies & jdk 8, spark\\
\hline
C8 & If available Link to developer documentation/manual & https://nlp.johnsnowlabs.com/api/ \\
\hline
C9 & Support email for questions & info@johnsnowlabs.com \\
\hline
\end{tabular}
\caption{Code metadata (mandatory)}
\label{} 
\end{table}

\section*{Current executable software version}
\label{exec}

\begin{table}[ht]
\begin{tabular}{|l|p{6.5cm}|p{6.5cm}|}
\hline
\textbf{Nr.} & \textbf{(Executable) software metadata description} & \textbf{Please fill in this column} \\
\hline
S1 & Current software version & 2.7.1 \\
\hline
S2 & Permanent link to executables of this version  & https://github.com/JohnSnowLabs/spark-nl \\
\hline
S3  & Permanent link to Reproducible Capsule & https://github.com/JohnSnowLabs/spark-nlp-workshop\\
\hline
S4 & Legal Software License & Apache-2.0 License  \\
\hline
S5 & Computing platforms/Operating Systems & Linux, Ubuntu, OSX, Microsoft Windows, Unix-like\\
\hline
S6 & Installation requirements \& dependencies & jdk 8, spark \\
\hline
S7 & If available, link to user manual - if formally published include a reference to the publication in the reference list & https://nlp.johnsnowlabs.com/api/ \\
\hline
S8 & Support email for questions &  info@johnsnowlabs.com\\
\hline
\end{tabular}

\end{table}

\end{document}